\documentclass{article}
\usepackage{ijcai25}

\usepackage{times}
\usepackage{soul}
\usepackage{url}
\usepackage[hidelinks]{hyperref}
\usepackage[utf8]{inputenc}
\usepackage[small]{caption}
\usepackage{graphicx}
\usepackage{amsmath}
\usepackage{amsthm}
\usepackage{booktabs}
\usepackage{algorithm}
\usepackage{algorithmic}
\usepackage[switch]{lineno}

\usepackage{pgfplots}
\usepackage{tikz}
\usepackage{amsfonts}
\usepackage{booktabs}
\usepackage{multirow}
\usepackage{xcolor}
\usepackage{colortbl}
\usepackage{amsfonts}
\usepackage{makecell}
\usepackage{pifont}
\usepackage{enumitem}

\usetikzlibrary{pgfplots.groupplots}
\pgfplotsset{compat=1.3}
\usetikzlibrary{patterns}
\definecolor{battleshipgrey}{rgb}{0.3, 0.3, 0.3}
\definecolor{brilliantrose}{rgb}{1.0, 0.33, 0.64}
\definecolor{americanrose}{rgb}{1.0, 0.01, 0.24}
\definecolor{jweigreen}{rgb}{0,0.45,0.24}
\definecolor{bluegray}{rgb}{0.1, 0.1, 0.4}
\definecolor{ao(english)}{rgb}{0.0, 0.5, 0.0}
\definecolor{blanchedalmond}{rgb}{1.0, 0.92, 0.8}
\definecolor{atomictangerine}{rgb}{1.0, 0.6, 0.4}
\definecolor{chocolate(web)}{rgb}{0.82, 0.41, 0.12}
\definecolor{bananayellow}{rgb}{1.0, 0.88, 0.21}
\definecolor{goldenbrown}{rgb}{0.6, 0.4, 0.08}
\definecolor{aliceblue}{rgb}{0.94, 0.97, 1.0}
\definecolor{beige}{rgb}{0.96, 0.96, 0.86}
\definecolor{babyblue}{rgb}{0.54, 0.81, 0.94}
\definecolor{camel}{rgb}{0.76, 0.6, 0.42}
\definecolor{cinnamon}{rgb}{0.82, 0.41, 0.12}
\definecolor{myblue}{RGB}{134, 181, 214}
\definecolor{myorange}{RGB}{255, 161, 78}

\title{Boosting Visual Knowledge-Intensive Training for LVLMs Through Causality-Driven Visual Object Completion}

\author{
Qingguo Hu$^{1,4}$\thanks{Equal contribution.}\and
Ante Wang$^{1}$\footnotemark[1]\and
Jia Song$^{1}$\and
Delai Qiu$^2$\and
Qingsong Liu$^2$\And
Jinsong Su$^{1,3,4}$\thanks{Corresponding author.}\\
\affiliations
$^1$School of Informatics, Xiamen University, China\\
$^2$Xiamen Unisound Intelligence Technology Co., Ltd\\
$^3$Shanghai Artificial Intelligence Laboratory, China\\
$^4$Key Laboratory of Digital Protection and Intelligent Processing of Intangible Cultural Heritage of Fujian and Taiwan (Xiamen University), Ministry of Culture and Tourism, China\\
\emails
\{huqingguo, wangante\}@stu.xmu.edu.cn,
jssu@xmu.edu.cn
}

\begin{document}

\maketitle

\begin{abstract}

Large Vision-Language Models (LVLMs) have experienced significant advancements in recent years. However, their performance still falls short in tasks requiring deep visual perception, such as identifying subtle differences between images. A potential cause is the scarcity of visual knowledge in popular instruction-tuning corpora, resulting in inadequate visual perception and reasoning capabilities.
To address this challenge, we introduce a self-improvement framework grounded in a novel visual knowledge-intensive task, \underline{C}ausality-driven \underline{V}isual object \underline{C}ompletion (CVC). This task requires LVLMs to infer the masked object in an image based on its \textit{causal} relationships with the other visible information. We first obtain rich examples cheaply through our automated instance construction pipeline, without relying on sophisticated LVLMs (\textit{e.g.}, GPT-4V) or human assistance. Then, LVLMs effectively self-improve through trial and error learning using these created instances.
Our experiments demonstrate substantial gains across four challenging specialized tasks and four widely-used comprehensive benchmarks. Especially on specialized tasks, our method achieves an average improvement of 5.4\% and 4.0\% compared to the corresponding baselines when utilizing LLaVA-1.5-7B and LLaVA-1.5-13B, respectively. The code is available at \url{https://github.com/XMUDeepLIT/CVC}.
\end{abstract}
\section{Introduction}
\begin{figure}[t]
    \centering
    \includegraphics[width=\linewidth, trim={0.03in 2.03in 0.03in 0in}, clip]{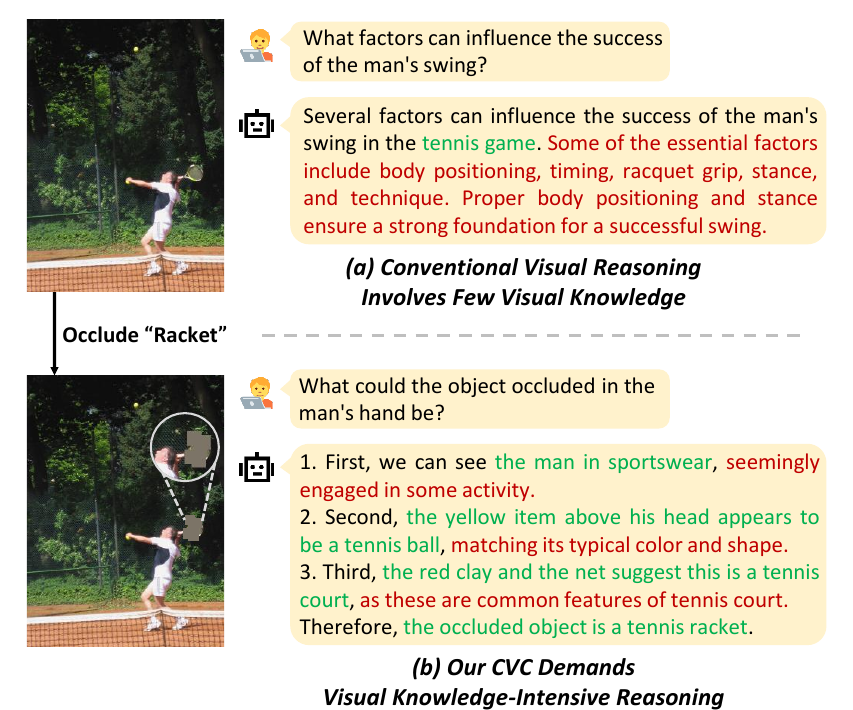}
    \caption{Comparison between conventional and our CVC instances for visual reasoning, where contents involving visual knowledge and linguistic commonsense knowledge are highlighted in green and red, respectively.}
    \label{fig:task}
\end{figure}

In the pursuit of Artificial General Intelligence (AGI), the ability to effectively process and understand multimodal information is of paramount importance. The surge in Large Language Models (LLMs)~\cite{achiam2023gpt} has catalyzed the development of powerful Large Vision-Language Models (LVLMs)~\cite{yang2023dawn,team2023gemini}. The main idea of most LVLMs is to integrate a pretrained visual encoder~\cite{radford2021learning} with LLMs using an alignment module~\cite{liu2023visual}. By leveraging the inherent capabilities of LLMs, these LVLMs have achieved excellent performance in various visual tasks, such as image captioning~\cite{yang2023dawn}, visual question answering~\cite{lan2024avg,liu2024improved}, and multimodal machine translation~\cite{lin2020dynamic,lan2023exploring,yin2023multi,lan2024translatotron}.

Despite the remarkable progress made by LVLMs, they still struggle with some basic visual perception and reasoning tasks, which humans can solve almost unerringly~\cite{tong2024eyes,fu2024blink}.
For instance,~\cite{tong2024eyes} demonstrate that existing LVLMs may perceive images with clear visual differences as similar, thus failing to distinguish them. 
This issue may stem from the inherent deficiency of current LVLM training corpora, which prioritize commonsense knowledge over visual knowledge and lack complex reasoning tasks involving visual scenarios.
In Figure~\ref{fig:task}(a), we showcase a typical example categorized as ``\textit{complex reasoning}'' within the popular LLaVA training corpus~\cite{liu2023visual}. The overall reasoning process involves limited visual knowledge (``\textit{playing tennis}''), but instead, mostly relies on the LLM's intrinsic commonsense knowledge (``\textit{the factors influencing the success of swing}'').
This can be because widely-used corpora~\cite{zhao2023svit,liu2023visual} are mostly crafted from strong language-only GPT-4~\cite{achiam2023gpt}, thus naturally contain a large amount of linguistic knowledge.
As a result, training with these instances only may limit the exploitation of LVLMs' capabilities in visual perception and reasoning.

To tackle this issue, we propose a novel self-improvement framework for further exploiting the visual capabilities of LVLMs autonomously.
Specially, in addition to existing vision-language tasks~\cite{goyal2017making,krishna2017visual,kang2023bigvideo}, we introduce causality-driven visual object completion (CVC), a challenging visual knowledge-intensive reasoning task for multimodal instruction tuning.
This task is inspired by ``\textit{visual completion}''~\cite{pessoa1998finding} in perceptual psychology, where humans with high-level cognitive processing and reasoning skills are capable of extracting meaning even from incomplete visual information.
Taking Figure~\ref{fig:task}(b) for example, given an image where an object is masked, the LVLM has to take the visible context as evidence and provide a step-by-step reasoning path (rationale) for explicitly inferring the masked object.
Similar ideas have been successfully applied to the fields of vision-language for self-supervised pretraining~\cite{chen2020uniter}. 
However, unlike these studies where the masked area is randomly selected, we especially emphasize the \textit{high causality} between the masked object and its surroundings.
This prevents the LVLM from forcibly fitting training targets that are uncertain and difficult to infer, encouraging it to conduct reasonable inference over the masked image.

To cost-effectively produce training instances of CVC, we first leverage widely available image-caption pairs~\cite{lin2014microsoft} and occlude the high-causality objects recognized in images.
Particularly, the causality score between an object and its image context is empirically estimated by the confidence of a Masked Language Model (\textit{e.g.}, RoBERTa~\cite{liu2019roberta}) on its corresponding caption entity.
With the crafted CVC instances above, we aim to synthesize valid rationales that lead to the target answer for LVLM training. To address this challenge, we apply \textit{trial and error learning}~\cite{young2009learning} to LVLMs for self-improvement: 
for each CVC instance, we first ask an LVLM to synthesize multiple rationales (trials) for inferring the masked object.
Then, the difficulty of each instance is assessed based on the frequency of its trials yielding the target answer. We select challenging instances that are valuable for training to enhance the LVLM's learning efficiency.
Finally, these successful self-generated trials are fed to the LVLM for self-improvement.
In this way, the LVLM is not only taught to recognize the detailed information of images (visual perception), but also encouraged to conduct ``\textit{slow thinking}'' \cite{daniel2017thinking} for explicitly predicting the masked object by leveraging its relevant surroundings (visual reasoning).
Thus, the visual capabilities of the LVLM can be comprehensively improved without the help of humans or sophisticated LVLMs (\textit{e.g.}, GPT-4V \cite{achiam2023gpt}).

To demonstrate the effectiveness of our proposed self-improvement framework, we conduct extensive experiments on challenging specialized tasks, including MMVP~\cite{tong2024eyes}, Winoground~\cite{thrush2022winoground}, \emph{V}$^*$Bench~\cite{wu2024v}, VSR~\cite{liu2023visual} and comprehensive benchmarks: MME~\cite{fu2023mme}, MMBench~\cite{liu2023mmbench}, SEEDBench~\cite{li2023seed} and MM-Vet~\cite{yu2023mm}. The results on LLaVA-1.5 family LVLMs show that our method consistently outperforms the corresponding baselines, particularly on the more challenging tasks of MMVP and Winoground, with improvements of +10.0\% and +8.2\%, respectively. Detailed analyses also indicate that our method can scale to larger data volumes, validating the promise of our method for more pervasive use.

\begin{figure*}[t]
    \centering
    \includegraphics[width=\linewidth, trim={0in 2.96in 0.08in 0.00in}, clip]{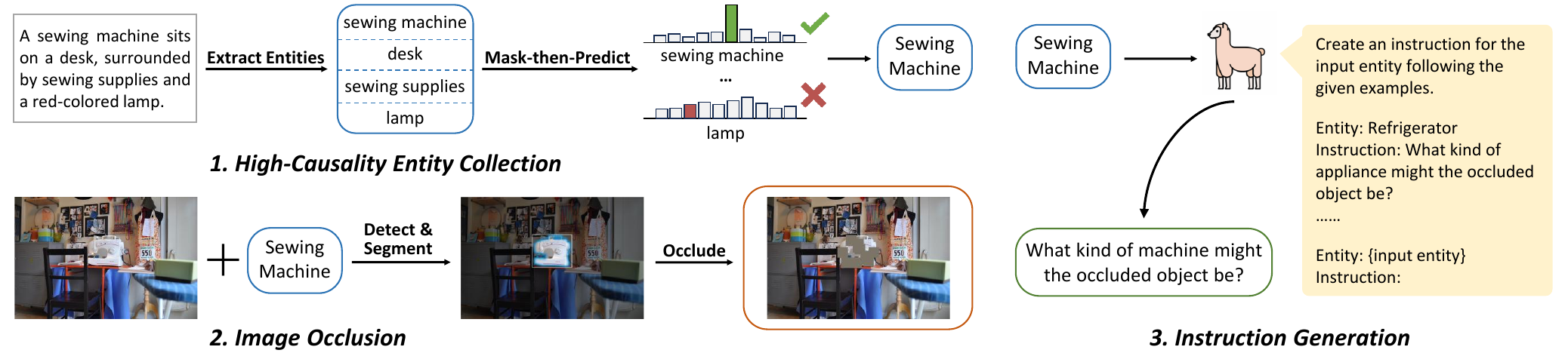}
    \caption{Overview of our data construction pipeline. Using widely available image-caption pairs, we construct CVC instances cost-effectively. Each CVC instance consists of a high-causality entity, an occluded image, and a task instruction.
    }
    \label{fig:data-preparation}
\end{figure*}

\section{Related Work}
\paragraph{Large Vision-Language Models (LVLMs).}
Mainstream LVLMs adopt a similar architecture~\cite{liu2023visual}, where a vision encoder is linked to an LLM via an alignment module, enabling perception of visual information. Nowadays, LVLMs have shown excellent performance in downstream tasks. However, recent studies highlight their limitations in visual understanding. For instance, \cite{tong2024eyes} benchmark LVLMs on distinguishing fine-grained visual differences, revealing the poor performance of existing LVLMs.
Similarly, \cite{fu2024blink} point out that existing benchmarks overlook visual perception, and instead use classic vision tasks to assess LVLMs. Though easy for humans, these tasks remain challenging for current models.

\paragraph{Learning from Rationales.}
Early studies have demonstrated that human-annotated rationales can enhance model performance~\cite{zhang2023multimodal}. Today, thanks to the emerging reasoning abilities in LLMs~\cite{wei2022chain,wang2025litesearch}, many studies~\cite{ho2022large,hsieh2023distilling} apply knowledge distillation to learn from synthetic rationales generated by advanced LLMs with hundreds of billions of parameters. However, the above approaches can be costly.
Therefore, an alternative line of research focuses on self-improvement methodologies, enabling LLMs to learn from self-generated rationales~\cite{huang2022large}.
In the field of LVLMs, such self-improvement techniques remain under-explored. In this paper, we harness LVLMs' inherent reasoning abilities to autonomously generate rationales, using them to cheaply enhance the LVLM’s visual capabilities.


\paragraph{Mask-then-Predict.}
This paradigm aims to enhance model performance by recovering masked signals. It has been extensively investigated in both language~\cite{devlin2018bert,liu2019roberta} and vision~\cite{bao2021beit,he2022masked} domains separately. Along this line, a series of works~\cite{chen2020uniter,kwon2022masked} employ this paradigm to conduct Vision-Language Pre-training (VLP).
Unlike conventional approaches in VLP where masked signals are directly predicted, our method requires the LVLM to infer the masked object through a step-by-step rationale. Additionally, we emphasize the high causality between the masked object and its surroundings, rather than applying random masking. By doing so, our method can further develop the LVLM's ability to utilize visible context and conduct reasonable inference over the masked image.

\section{Our Framework}


In this section, we introduce an innovative self-improvement framework for LVLMs. We start by providing a detailed formal definition of causality-driven visual object completion (CVC), a challenging visual knowledge-intensive reasoning task for LVLMs (\S \ref{sec:task}). Next, we present an overview of the pipeline used to construct task instances cheaply (\S \ref{sec:data-preparation}). Finally, using the constructed CVC instances, we utilize an LVLM to autonomously synthesize valid rationales. These self-generated data are used as additional training data for the multimodal instruction tuning of the LVLM (\S \ref{sec:training}).


\subsection{Task Definition}
\label{sec:task}

Formally, given an occluded image ${I_{\backslash e}}$, where an object ${e}$\footnote{For the sake of clarity, we denote both an object in an image and its corresponding entity by $e$.} having high causality with its surroundings is artificially masked, we ask an LVLM to infer the masked object by providing a step-by-step rationale $r$:
\begin{equation}
r = \mathtt{LVLM}({I_{\backslash e}}, {q}, {p}),
\end{equation}
where the rationale $r$ refers to natural language explanations that support the model’s final prediction, $q$ is the task instruction (\textit{e.g.}, \textit{``What is the occluded object?''}), and $p$ is the Chain-of-Thought (CoT) prompt (\textit{i.e.}, \textit{``Let’s think step by step''}) used to elicit the rationale. In this context, high causality refers to the strong logical association between the masked object and its surroundings, enabling reasonable inference.

Compared with predicting the object $e$ only, providing a step-by-step rationale $r$ can better help the LVLM learn to search for relevant cues in the visible context and then conduct causal reasoning over them to infer the answer.
Therefore, deep visual perception and reasoning capabilities of the LVLM can be promoted through this process.

\subsection{Data Preparation}
\label{sec:data-preparation}

Based on widely available image-caption pairs, we develop an automatic data construction pipeline to create the dataset of CVC, where each instance is composed of $\{e, I_{\backslash e}, q\}$.
As shown in Figure~\ref{fig:data-preparation}, we successively construct the three elements of each instance through
\textit{(i) high-causality entity collection}, \textit{(ii) image occlusion}, and \textit{(iii) instruction generation}. 
We describe these steps in detail below.\footnote{Further implementation details are provided in \autoref{sec:appendix:implement}.}

\paragraph{High-Causality Entity Collection.}
Directly estimating the causality between an object and its image context is non-trivial.
Therefore, we empirically compute the causality score of an object by utilizing the uncertainty estimation of the corresponding entity in the caption, which shares the same semantic meaning as the corresponding image.
Specifically, given an image-caption pair $ \{ I, T \}$, where $I$ denotes the image and $T$ represents the caption, we first employ LLaMA2-7B~\cite{touvron2023llama2} to extract entities (\textit{e.g}., $e_i$) from $T$ via in-context learning. 
Next, we mask an entity $e_i$ in the caption $T$, yielding the masked text $T_{\backslash e_i}$.
We then feed $T_{\backslash e_i}$ into RoBERTa~\cite{liu2019roberta} (denoted as $\phi$) to predict the masked tokens.
The prediction probability of $e_i$ serves as the causality score: $p( e_i \mid T_{\backslash e_i}  ; \phi)$.
Intuitively, entities with high causality are more easily predicted based on their context, resulting in higher probabilities compared to low-causality entities.
Finally, we select those high-causality entities with scores exceeding a threshold $\gamma$.
\paragraph{Image Occlusion.}
After collecting high-causality entities, we proceed to occlude these entities in their corresponding images. First, we utilize GLIP~\cite{li2022grounded} to ground each entity’s corresponding object in the image, thereby yielding a bounding box. This bounding box serves as input for SAM~\cite{kirillov2023segment}, which predicts a mask that precisely delineates the region of the object. We then apply heavy occlusion to the segmented pixels.
We use rectangular boxes instead of actual contours to occlude objects.
This prevents the LVLM from relying on shapes for predictions thus hindering its effective use of surroundings for reasoning.



\paragraph{Instruction Generation.}
The task instruction aims to elicit the desired LVLM predictions.
Given the diverse range of entities we collected, a fixed instruction (\textit{e.g.}, ``\textit{What is the occluded object?}'') might lead to ambiguous references, 
thereby reducing the data efficiency because it is hard for an LVLM to yield the answers exactly.
Besides, it limits the data diversity, which can be crucial for model generalization.
Therefore, we construct a specialized instruction for each entity. To achieve this, we manually craft several examples of entity-instruction pairs and use LLaMA2-7B through in-context learning to generate a new instruction based on the input entity.
\begin{figure}[t]
    \centering
    \includegraphics[width=\linewidth, trim={0in 1.29in 7.10in 0.79in}, clip]{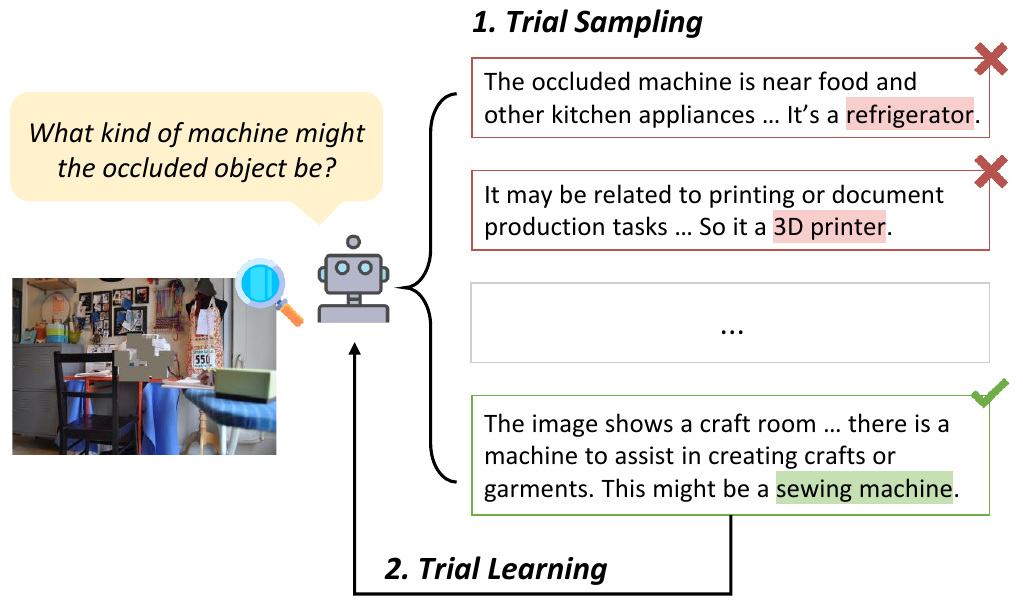}
    \caption{Overview of our self-improvement approach. Based on trial and error learning, the LVLM samples multiple rationales (trials) for a given CVC instance. The sampled trials are selectively learned based on their correctness and estimated difficulty.  }
    \label{fig:self-improvement}
\end{figure}

\subsection{Model Training}
\label{sec:training}

Thus far, the crafted CVC instances still lacks valid rationales that lead to the target entity. To address this, we apply \textit{trial and error learning}~\cite{young2009learning}, a core learning mechanism in behavioral science, to LVLMs for self-improvement:
faced with an unfamiliar problem, the learner experiments with multiple trials, evaluates their successfulness based on environmental feedback, and learns from successful attempts to improve problem-solving capabilities.
As shown in Figure~\ref{fig:self-improvement}, we illustrate how successful trials are sampled and  then fed back into the LVLM for self-improvement.


\paragraph{Trial Sampling.}
Given each CVC instance $\{e, I_{\backslash e}, q\}$ and an LVLM parameterized by $\theta$,
each rationale produced by the LVLM for inferring the masked object is treated as a trial. 
We adopt the popular nucleus sampling~\cite{holtzman2019curious} to obtain $N$ different trials:
\begin{equation}
r_1, r_2, \ldots,  r_N \sim \mathtt{LVLM}({I_{\backslash e}}, {q}, {p}).
\end{equation}

Then, we assess the successfulness of each trial $r_j$ by verifying its corresponding answer $\hat{e}_j$ against the target entity $e$. 
Concretely, for each trial $r_j$, we extract the answer $\hat{e}_j$ utilizing LLaMA2-7B through in-context learning\footnote{We provide the prompting examples in \autoref{sec:appendix:prompt}.}, and $r_j$ is successful when $\hat{e}_j = e$.


\paragraph{Trial Learning.}
With the obtained successful trials for each instance, we only select challenging ones that are worth learning to train the LVLM for improving learning efficiency.
To achieve this, we measure the difficulty of a CVC instance by calculating the inverse frequency of its trials that yield the target answer $e$: $\mathcal{F} = 1 - \frac{1}{N}\sum_{k=1}^N \mathbb{I}(\hat{e}_k = e)$, where $\mathbb{I}(\cdot)$ denotes the indicator function.
Then, only successful trials from instances with difficulty scores higher than the threshold $\alpha$ are chosen.
Finally, for an instance $\{e, I_{\backslash e}, q \}$ with a successful trial $r'$, we train the LVLM by hybridizing the trial and target entity as supervised signals, using the following standard cross-entropy loss:
\begin{equation}
    \mathcal{L} = - \left( \log p(e \mid I_{\backslash e}, q ; \theta) + \log p(r' \mid I_{\backslash e}, q, p ; \theta) \right).
\end{equation}
The hybrid loss term promotes the ``\textit{fast and slow thinking}'' of the LVLM via learning to not only predict the direct answer but also decompose the complex visual completion process step by step. To maintain the LVLM's instruction-following ability, we combine our data with general multimodal instruction data and jointly perform multimodal instruction tuning.

Through this process, without the need for sophisticated LVLMs (\textit{e.g.}, GPT-4V) or human annotations, the LVLM autonomously learns from its own scarce but successful trials. While achieving proficiency in CVC, the LVLM's visual perception and reasoning capabilities are further developed.

\section{Experiments}
\begin{table*}[t]
    \centering
    \scalebox{0.9}{
    \begin{tabular}{lcccccccc}
    
    \toprule
    & \multicolumn{4}{c}{Specialized Task}& \multicolumn{4}{c}{Comprehensive Benchmark} \\
    
    \cmidrule(lr){2-5}\cmidrule(lr){6-9}

    Method & MMVP & Winoground  & \emph{V}$^*$Bench & VSR & MME & MMBench & SEEDBench & MM-Vet   \\

    \midrule

    LLaVA~\cite{liu2023visual} & 6.0 & - & 35.6 & - & 809.6 & 38.7 & 37.0 & 25.5 \\
    InstructBLIP~\cite{dai2023instructblipgeneralpurposevisionlanguagemodels} & 16.7 & - & 34.0 & - & 1212.8 & 36.0 & 58.8 & 26.2 \\
    Gemini Pro~\cite{team2023gemini} & 40.7 & - & 48.2 & - & 1496.6 & 73.6 & 62.4 & 64.3   \\
    GPT-4V~\cite{achiam2023gpt} & 38.7 & - & 55.0 & - & 1409.4 & 81.0 & 69.1 & 67.7   \\
  
    \midrule
    Naive-VC & 27.3 & 29.5 & 48.7 & 66.9 & 1473.8 & 65.6 & 66.3 & 30.8 \\
    \midrule
    
    LLaVA-1.5-7B~\cite{liu2024improved} & 20.7 & 25.3 & 47.6 & 66.9 & 1510.7 & 64.3 & 66.1 & 31.1    \\
    \textbf{w/ CVC} & \textbf{30.7} & \textbf{33.5} & \textbf{49.7} & \textbf{68.1} & \textbf{1519.5} & \textbf{66.6} & \textbf{67.1}  & \textbf{34.1}  \\
     \rowcolor{gray!20} $\Delta$ & +10.0 & +8.2 & +2.1 & +1.2 & +8.8 & +2.3 & +1.0 & +3.0 \\

     \midrule
    
    LLaVA-1.5-13B~\cite{liu2024improved}  & 33.3 & 34.3 & 48.7 & 67.3 & 1531.3 & 67.7 & 68.2 & 36.1   \\
    \textbf{w/ CVC} & \textbf{36.0} & \textbf{39.0} & \textbf{51.8} & \textbf{72.7} & \textbf{1543.8} & \textbf{70.1} & 68.2 & \textbf{38.2}  \\
     \rowcolor{gray!20} $\Delta$  & +2.7 & +4.7 & +3.1 & +5.4 & +12.5 & +2.4 & +0.0 & +2.1 \\
    \bottomrule
    \end{tabular}
    
    }
    \caption[]{Performance of our method on LLaVA-1.5 across all evaluation datasets. Baseline results are primarily sourced from~\cite{tong2024eyes},~\cite{wu2024v} and~\cite{liu2024improved}. Results for GPT-4V and Gemini Pro on comprehensive benchmarks are obtained from the respective official leaderboards.\protect\footnotemark}
    \label{tab:main-results}
\end{table*}

\subsection{Setup}

\paragraph{Evaluation Datasets.}
We conduct in-depth analyses on a range of challenging specialized tasks and widely-used comprehensive benchmarks, aiming to test the effectiveness of our method on the deep visual perception and general capabilities of LVLMs, respectively.
\begin{itemize}
    \item \textbf{Challenging specialized tasks:} 
    MMVP~\cite{tong2024eyes}, Winoground~\cite{thrush2022winoground}, \emph{V}$^*$Bench~\cite{wu2024v}, and VSR~\cite{liu2023visualspatial}. Among these, MMVP focuses on distinguishing fine-grained visual differences that LVLMs often overlook. Winoground tasks LVLMs to select the correct caption for an image from two options with easily confused spatial relationships. \emph{V}$^*$Bench challenges the model to recognize subtle visual details, and VSR assesses visual spatial understanding capability.
    \item \textbf{Comprehensive benchmarks:} 
    MME~\cite{fu2023mme}, MMBench~\cite{liu2023mmbench}, SEEDBench~\cite{li2023seed}, and MM-Vet~\cite{yu2023mm}. These benchmarks encompass a wide range of subtasks, providing a thorough assessment of our method's generalization.
\end{itemize}
We follow \cite{liu2024improved} to use the same testing scripts and evaluation metrics for fair comparison.

\paragraph{Implementation Details.}
Since most mainstream LVLMs share the same architecture, we follow~\cite{zhou2024calibrated} to choose popular LLaVA-1.5~\cite{liu2024improved} for experiments.
Both the 7B and 13B versions are used for demonstrating the scalability of our method across various model sizes. 
The CVC instances are constructed based on COCO dataset~\cite{lin2014microsoft}. We set $\gamma$, $N$, and $\alpha$ to 0.3, 16, and 0.75, respectively.
By default, we use 90K of our data for training across all experiments unless otherwise noted. During training, we combine our data with the 665K instruction data from LLaVA-1.5 for multimodal instruction tuning. To ensure a fair comparison, our training starts from the pretrained (\textit{i.e.}, not yet instruction-tuned) weights of LLaVA-1.5, following the same training hyperparameters. All experiments are conducted on 8 × A100 80G GPUs.

\begin{table}[t]
    \centering
    \scalebox{0.9}{
    \begin{tabular}{lcccccc}
    
    \toprule
    & \multicolumn{3}{c}{MMBench} & \multicolumn{3}{c}{SEEDBench} \\

    \cmidrule(lr){2-4}\cmidrule(lr){5-7}
    Method & $\rm{LR}$ & $\rm{RR}$ & $\rm{AR}$ & $\rm{VR}$ & $\rm{AP}$  & $\rm{II}$ \\
    
    \midrule
     LLaVA-1.5-13B & 44.1 & 62.6 & 70.4 & 77.0 & 38.6 & 73.2 \\
     \midrule
    
    LLaVA-1.5-7B & 30.5 & 53.0 & \textbf{73.4} & 76.7 & 33.7 & 69.1 \\
    \textbf{w/ CVC} & \textbf{33.1} & \textbf{64.4} & 70.4 & \textbf{77.6} & \textbf{37.0} & \textbf{70.1}\\
    \rowcolor{gray!20} $\Delta$ & +2.6 & +11.4 & -3.0 & +0.9 & +3.3 & +1.0 \\
   
    \bottomrule
    \end{tabular}
    }
    \caption{Results on visual reasoning tasks. The abbreviations for these tasks are as follows: Logical Reasoning (LR), Relation Reasoning (RR), and Attribute Reasoning (AR) for MMBench; and Visual Reasoning (VR), Action Prediction (AP), and Instance Interaction (II) for SEEDBench.}
    \label{tab:cognitive-reasoning}
\end{table}

\paragraph{Baselines.} To better validate the effect of our framework,
we introduce a baseline termed Naive-VC. 
It is based on LLaVA-1.5-7B and trained extensively on data of the naive visual completion task, where entities are \textit{randomly} selected before masking the corresponding objects in images.
Additionally, we include results from other open-source LVLMs like LLaVA~\cite{liu2023visual} and InstructBLIP~\cite{dai2023instructblipgeneralpurposevisionlanguagemodels}, as well as state-of-the-art models like GPT-4V~\cite{achiam2023gpt} and Gemini Pro~\cite{team2023gemini}, to further demonstrate the effectiveness of our method.

\begin{table*}[t]
    \centering
    \scalebox{0.9}{
    \begin{tabular}{lccccccccc}
    
    \toprule
    & & \multicolumn{4}{c}{Specialized Task}& \multicolumn{4}{c}{Comprehensive Benchmark} \\
    
    \cmidrule(lr){3-6}\cmidrule(lr){7-10}

    Synthesizer & Recall ($\%$) & MMVP & Winoground  & \emph{V}$^*$Bench & VSR & MME & MMBench & SEEDBench & MM-Vet   \\



    
    \midrule

    MiniGPT-v2 & \textbf{17.3} & \textbf{32.7} & \underline{32.3}  & \textbf{52.9} & \underline{68.7} & \underline{1516.5} & 65.8 & \underline{67.4} & 32.7 \\
    LLaVA-1.5-7B & \underline{12.1} & \underline{30.7} & \textbf{33.5} & \underline{49.7} & 68.1 & \textbf{1519.5} & \textbf{66.6} & 67.1  & \textbf{34.1} \\
    LLaVA-NeXT-34B & 10.3 & \underline{30.7} & 31.8  & 48.7 & \textbf{70.8} & 1497.8 & \underline{66.3} & \textbf{67.6} & \underline{33.2} \\

    \bottomrule
    \end{tabular}
    }
    \caption{Comparison of different LVLMs used as the rationale synthesizer. All training is conducted on LLaVA-1.5-7B. Recall indicates the proportion of successful rationales generated by each synthesizer. The best results are bolded, and the second-best results are underlined.}
    \label{tab:diffenert-LVLMs}
\end{table*}

\subsection{Main Results}

\footnotetext{We provide the URL for the leaderboard of each comprehensive benchmark in \autoref{sec:appendix:leaderboard}.}

\paragraph{\textit{Our Method Notably Boosts Deep Visual Perception.}}
As shown in Table~\ref{tab:main-results}, our method achieves notable improvements on 4 challenging specialized tasks. 
For LLaVA-1.5-7B, our method achieves nearly a 10\% improvement on the two difficult tasks of Winoground and MMVP. 
Consistent performance gains are also significant in \emph{V}$^*$Bench and VSR. 
Compared with Naive-VC, our method achieves more pronounced improvements.
It is because CVC encourages the LVLM to engage in reasonable inference by leveraging richer contextual cues, demonstrating the importance of leveraging high-causality objects for visual completion.
Notably, LLaVA-1.5-7B equipped with CVC data achieves performance comparable to or even surpassing vanilla LLaVA-1.5-13B on these tasks. 
For LLaVA-1.5-13B, our method also achieves substantial performance gains, with improvements of nearly 2\%$\sim$5\%.
These results demonstrate that our method can effectively enhance this more advanced LVLM, showcasing its potential to improve well-developed models at larger scales.
Regarding comprehensive benchmarks, although our method is task-agnostic, it effectively enhances the performance of both LLaVA-1.5-7B and LLaVA-1.5-13B, achieving average improvements of 1.69\% and 1.28\%, respectively. 
Given that these tasks do not emphasize deep perception but involve a broader range of visual knowledge, these results indicate that our method effectively injects intensive visual knowledge into LVLMs and further aligns it within the language domain, demonstrating the generalizability of our approach.

\paragraph{\textit{Our Method Further Enhances Reasoning in Visual Scenarios.}}
Since our method also encourages the LVLM to engage in visual reasoning, we further investigate how it impacts the reasoning ability of the LVLM in visual scenarios. To this end, we assess performance on several reasoning subtasks from MMBench and SEEDBench, which require not only basic perception but also cognitive inference~\cite{li2023seed,liu2023mmbench}.
As shown in Table~\ref{tab:cognitive-reasoning}, the performance of LLaVA-1.5-7B is generally improved on these reasoning tasks. Notably, the performance on relation reasoning exhibits the most substantial improvement, with an increase of 11.4\%. These results indicate that our method effectively enhances the LVLM's reasoning ability, further validating its effectiveness from another perspective.


\pgfplotsset{compat=1.3,
    /pgfplots/ybar legend/.style={
    /pgfplots/legend image code/.code={%
       \draw[##1,/tikz/.cd,yshift=-0.25em]
        (0cm,0cm) rectangle (7pt,0.8em);},
   },
}
\begin{figure}[t]
\hspace*{-0.1cm}
\pgfplotsset{width=4cm, height=4.25cm}
    \centering
    \begin{tikzpicture}  
    \begin{groupplot}[
          group style={
          group name=plot,
          horizontal sep=20pt,
          vertical sep=20pt,
          group size=2 by 2},]
      \nextgroupplot[
            ybar,
            ymin=15, ymax=33,
            ytick={15, 20, 25, 30},
            major x tick style = transparent,
            bar width=6pt,
            enlarge x limits=0.25,
            ylabel={Accuracy (\%)},
            symbolic x coords={MMVP},  
            xtick=data,  
            y label style={at={(axis description cs:-0.25,0.5)},anchor=south},
            axis x line*=bottom,
            axis y line*=left,
            legend cell align=center,
            legend style={
                    at={(2.7,-0.5)},
                    anchor=south,
                    column sep=0.5ex,
                    font=\small,
                    draw=none,
                    legend columns=1,
                }
            ]  
        \addplot[ybar, fill=bananayellow,  mark=, postaction={}] coordinates {
            (MMVP, 20.7)
        };
        \addplot[ybar, fill=cinnamon,  postaction={}] coordinates {
            (MMVP, 26.0)
        };  
        \addplot[ybar, fill=cinnamon,  postaction={pattern=north east lines}] coordinates {
            (MMVP, 29.3)
        };  
        \addplot[ybar, fill=camel,  postaction={pattern=north west lines}] coordinates {
            (MMVP, 30.7)
        };  
        \addplot[ybar, fill=camel,  postaction={}] coordinates {
            (MMVP, 32.0)
        };
        \legend{
            0K, 
            30K,
            50K,
            90K,
            150K,
            }
        ]
      \nextgroupplot[
            ybar,
            ymin=20, ymax=38,
            ytick={20, 25, 30, 35},
            major x tick style = transparent,
            bar width=6pt,
            enlarge x limits=0.25,
            symbolic x coords={Winoground},  
            xtick=data,  
                axis x line*=bottom,
                axis y line*=left,
        legend cell align=left,
                legend style={
                        at={(1,1.05)},
                        anchor=south east,
                        column sep=1ex,
                        font=\small,
                }
            ]
        \addplot[ybar, fill=bananayellow,  mark=, postaction={}] coordinates {
            (Winoground, 25.3)
        };
        \addplot[ybar, fill=cinnamon,  postaction={}] coordinates {
            (Winoground, 29.5)
        };  
        \addplot[ybar, fill=cinnamon,  postaction={pattern=north east lines}] coordinates {
            (Winoground, 33.3)
        };  
        \addplot[ybar, fill=camel,  postaction={pattern=north west lines}] coordinates {
            (Winoground, 33.5)
        };  
        \addplot[ybar, fill=camel,  postaction={}] coordinates {
            (Winoground, 36.3)
        };  
        \nextgroupplot[
            ybar,
            ymin=45, ymax=52.1,
            ytick={45, 47, 49, 51},
            ylabel={Accuracy (\%)},
            major x tick style = transparent,
            bar width=6pt,
            enlarge x limits=0.25,
            symbolic x coords={\emph{V}$^*$Bench},  
            xtick=data,  
                axis x line*=bottom,
                axis y line*=left,
        legend cell align=left,
                legend style={
                        at={(1,1.05)},
                        anchor=south east,
                        column sep=1ex,
                        font=\small,
                }
            ]  
        \addplot[ybar, fill=bananayellow,  mark=, postaction={}] coordinates {
            (\emph{V}$^*$Bench, 47.6)
        };
        \addplot[ybar, fill=cinnamon,  postaction={}] coordinates {
            (\emph{V}$^*$Bench, 49.7)
        };  
        \addplot[ybar, fill=cinnamon,  postaction={pattern=north east lines}] coordinates {
            (\emph{V}$^*$Bench, 48.7)
        };  
        \addplot[ybar, fill=camel,  postaction={pattern=north west lines}] coordinates {
            (\emph{V}$^*$Bench, 49.7)
        };  
        \addplot[ybar, fill=camel,  postaction={}] coordinates {
            (\emph{V}$^*$Bench, 50.8)
        };
        \nextgroupplot[
            ybar,
            ymin=65, ymax=67.5,
            ytick={65, 66, 67},
            major x tick style = transparent,
            bar width=6pt,
            enlarge x limits=0.25,
            symbolic x coords={SEEDBench},  
            xtick=data,  
                axis x line*=bottom,
                axis y line*=left,
        legend cell align=left,
                legend style={
                        at={(1,1.05)},
                        anchor=south east,
                        column sep=1ex,
                        font=\small,
                }
            ]  
        \addplot[ybar, fill=bananayellow,  mark=, postaction={}] coordinates {
            (SEEDBench, 66.1)
        };
        \addplot[ybar, fill=cinnamon,  postaction={}] coordinates {
            (SEEDBench, 66.9)
        };  
        \addplot[ybar, fill=cinnamon,  postaction={pattern=north east lines}] coordinates {
            (SEEDBench, 66.42)
        };  
        \addplot[ybar, fill=camel,  postaction={pattern=north west lines}] coordinates {
            (SEEDBench, 67.13)
        };  
        \addplot[ybar, fill=camel,  postaction={}] coordinates {
            (SEEDBench, 66.39)
        };
        ]
\end{groupplot}
    \end{tikzpicture}
    \caption{Results of scaling up CVC data for LLaVA-1.5-7B.}
    \label{fig:scalability}
\end{figure}
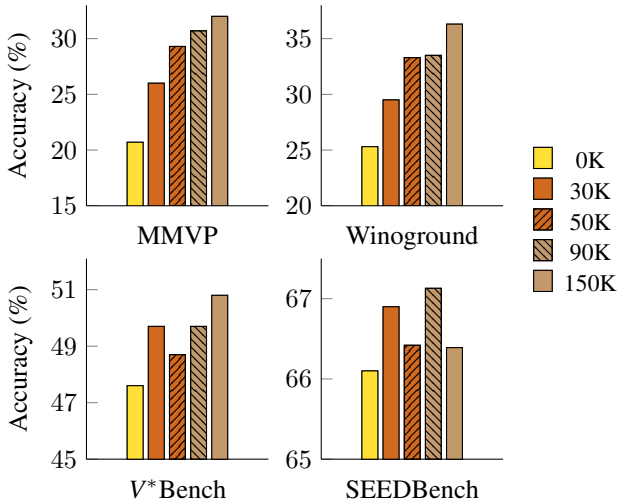

\subsection{Analysis}

\paragraph{\textit{Effect of Scaling Up CVC Data.}}
We further investigate the effect of CVC data scale on downstream visual tasks. We apply our method to LLaVA-1.5-7B with varying CVC data scales, ranging from 30\text{K} to 150\text{K}. To highlight the effectiveness of our method on the deep visual perception, we focus on three challenging tasks: MMVP, Winoground and \emph{V}$^*$Bench. The results are shown in Figure~\ref{fig:scalability}. We can observe a consistent trend of performance improvement across three tasks as the CVC data scale increases. Notably, after training with 150\text{K} of CVC data, our method achieves an accuracy of 36.3\% on Winoground and 32.0\% on MMVP, showing absolute improvements of 11.0\% and 11.3\% compared to the vanilla LLaVA-1.5-7B. A positive impact of CVC data scale on \emph{V}$^*$Bench is also observed, though it is relatively less pronounced.
Analyzing the effect of each data scale setting, we find that the 30\text{K} setting yields the most significant incremental improvement across the three tasks. After exceeding 90\text{K}, the rate of performance improvement slows, but the LVLM continues to benefit from additional data. However, in general scenarios (e.g., SEEDBench), we observe a slight performance decline when scaling up to 150\text{K}. This may be attributed to the excessive proportion of CVC data in the training corpus, which diminishes the impact of general data.
Nonetheless, from the perspective of core visual perception, these results demonstrate that our method is not only efficient with limited data but also scalable to larger data volumes.

\paragraph{\textit{Influence of Different Synthesizers.}}
So far, we have shown the effectiveness of our method in self-improvement. Nevertheless, other LVLMs can also function as the rationale synthesizer within our framework. In Table~\ref{tab:diffenert-LVLMs}, we present the performance of MiniGPT-v2~\cite{chen2023minigpt} and LLaVA-NeXT-34B~\cite{liu2024llavanext} as the synthesizer. Specifically, MiniGPT-v2 is an LVLM based on LLaMA2-7B, and LLaVA-NeXT-34B is a more advanced LVLM trained on superior multimodal data. The results first reveal that CVC presents a substantial challenge to current LVLMs, as evidenced by the extremely low recall of each synthesizer. Second, our results show that different synthesizers yield comparable improvements. We attribute this to our method ensuring comparable data quality across the synthesizers by validating rationales with gold answers. As a result, our method is still effective when applied to less sophisticated LVLMs. Third, the improvement of our method in deep visual perception is closely tied to the synthesizer’s performance on CVC. For example, MiniGPT-v2, with the highest recall of 17.3\%, exhibits overall more advanced performance on 4 challenging specialized tasks. In contrast, although LLaVA-NeXT-34B has achieved state-of-the-art results across a wide range of benchmarks~\cite{liu2024llavanext}, its lower proficiency in CVC results in relatively modest improvements. This suggests that utilizing LVLMs with higher proficiency in CVC could further enhance the potential of our method. We hope these findings offer valuable insights for future research.

\pgfplotsset{compat=1.3,
    /pgfplots/ybar legend/.style={
    /pgfplots/legend image code/.code={%
       \draw[##1,/tikz/.cd,yshift=-0.25em]
        (0cm,0cm) rectangle (7pt,0.8em);},
   },
}
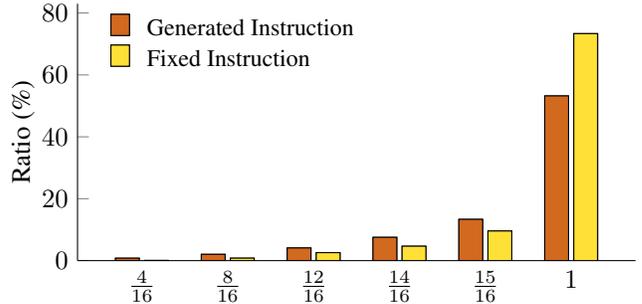
\begin{figure}[t]
\hspace*{-0.3cm}
\pgfplotsset{width=9cm, height=5cm}
    \centering
    \begin{tikzpicture}  
    \begin{axis}[
        ybar,
        ymin=0, ymax=83,
        ytick={0, 20, 40, 60, 80},
        major x tick style = transparent,
        bar width=9pt,
        enlarge x limits=0.15,
        ylabel={Ratio (\%)},
        symbolic x coords={$\frac{4}{16}$, $\frac{8}{16}$, $\frac{12}{16}$, $\frac{14}{16}$, $\frac{15}{16}$, $1$},
        xtick=data,
        xticklabel style={yshift=4pt},
        x label style={at={(axis description cs:0.5,-0.15), anchor=north}},
        y label style={at={(axis description cs:-0.06,0.5)},anchor=south},
        axis x line*=bottom,
        axis y line*=left,
        legend cell align=left,
        legend style={
                at={(0.52, 0.70)},
                anchor=south east,
                column sep=1ex,
                font=\small,
                draw=none,
                legend columns=2,
                transpose legend,
            }
        ]
        \addplot[ybar, fill=cinnamon, postaction={}] coordinates {
            ($1$, 53.21) ($\frac{15}{16}$, 13.34) ($\frac{14}{16}$, 7.53) ($\frac{12}{16}$, 4.09) ($\frac{8}{16}$, 2.04) ($\frac{4}{16}$, 0.79)
        };
        \addplot[ybar, fill=bananayellow, mark=, postaction={}] coordinates {
            ($1$, 73.34) ($\frac{15}{16}$, 9.57) ($\frac{14}{16}$, 4.66) ($\frac{12}{16}$, 2.57) ($\frac{8}{16}$, 0.80) ($\frac{4}{16}$, 0.07)
        };
        \legend{
        Generated Instruction,
        Fixed Instruction
            }
    \end{axis}
    \end{tikzpicture}
    \caption{Comparison between two different task instruction settings in the distribution of $\mathcal{F}$ across all CVC instances. Both settings use LLaVA-1.5-7B for rationale synthesis.}
    \label{fig:frequency-distribution}
\end{figure}
\begin{figure*}[t]
    \centering
    \includegraphics[width=\linewidth, trim={0.75in 1.58in 0.5in 0.0in}, clip]{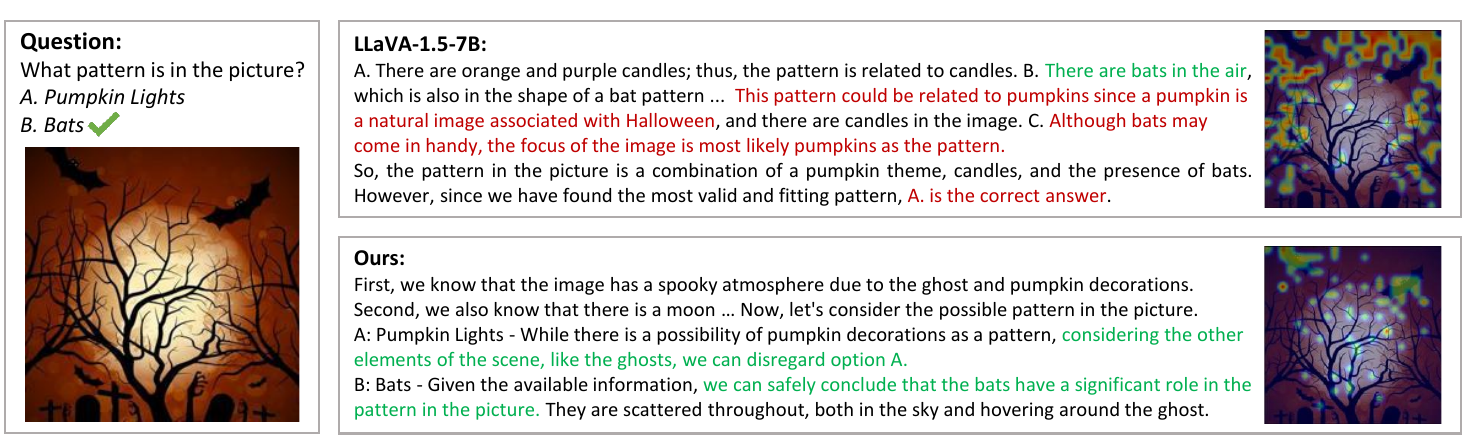}
    \caption{An example comparison of LLaVA-1.5-7B with and without our method. \textbf{Left}: A test sample from MMVP includes a question and an image, with the correct option annotated. \textbf{Right}: The reasoning processes and relative attention maps of the two methods. The highlighted regions in the attention map reveal where this method places more attention compared to the other.}
    \label{fig:case}
\end{figure*}

\paragraph{\textit{Generated Instruction Promotes Data Diversity.}}
To evaluate the effect of generated instruction, we compare this setting with using a fixed instruction (``\textit{What is the occluded object?}'').
First, our results demonstrate that using generated instructions significantly outperforms a fixed instruction in terms of data diversity. As shown in Figure~\ref{fig:frequency-distribution}, using a fixed instruction results in nearly 20\% more CVC instances with no successful rationale compared to using generated instructions. This indicates that generated instruction can lead to more specific references, thereby aiding the LVLM in producing accurate answers. Due to this design, our data covers nearly 2,000 different entities, ensuring data diversity. Second, in Table~\ref{tab:generated-fixed}, using generated instructions exhibits better performance on both specialized and general visual tasks. 

\paragraph{\textit{Harder CVC Instances Boost Greater Performance.}} 
We measure the difficulty of a CVC instance based on the frequency of its trials that yield the target answer. To investigate how the difficulty of CVC instances impacts LVLM performance, we collect CVC instances with different difficulties for LVLMs training. As depicted in Table~\ref{tab:generated-fixed}, the harder CVC instances (\( \alpha = 0.75\)) exhibit a more pronounced performance improvement compared to the easier ones (\( \alpha = 0.5\)). This finding underscores that training with harder CVC instances drives the LVLM to achieve more substantial gains in tackling complex visual scenarios, highlighting the importance of instance difficulty in boosting model performance.

\paragraph{\textit{Effect of Hybrid Loss.}}
Another key decision of our method is how to format the synthesized data for LVLMs training. To investigate this, we conduct an ablation study of the hybrid loss. As shown in Table~\ref{tab:hybrid-formats}, both rationale-only and direct-answer-only training can generally enhance the LVLM's performance on downstream visual tasks, with the exception of MM-Vet. When combining both rationale and direct answer for training, the hybrid loss enables the LVLM to engage in ``\textit{fast and slow thinking},'' leading to further improvements in overall performance.

    
    
     

    
    

\begin{table}[t]
    \centering
    \scalebox{0.9}{
    \begin{tabular}{l|cccc}
    
    \toprule
     \multirow{2}{*}{Method} & \multirow{2}{*}{MMVP} & Wino & {SEED} & \multirow{2}{*}{MM-Vet}\\
    
      & & ground & Bench &   \\
     
     \midrule

    LLaVA-1.5-7B & 20.7 & 25.3 & 66.1 & 31.1 \\
    w/ CVC (\(\alpha=0.5\)) & 30.0 & 29.5 & 66.9 & 31.7\\
    w/ CVC (\(\alpha=0.75\)) & \textbf{30.7} & \textbf{33.5} & \textbf{67.1} & \textbf{34.1}  \\
    \hspace{4mm} w/o Generated Inst. & 28.7 & 32.5 & 67.1 & 31.8 \\
    \bottomrule
    \end{tabular}
    
    }
    \caption{Results of our method enhanced by introducing generated instructions (Generated Inst.) and harder CVC instances. We conduct this ablation study on LLaVA-1.5-7B.}
    \label{tab:generated-fixed}
\end{table}

    
    


    
    

\subsection{Case Study}
We present a case study to underscore the efficacy of our method in visual reasoning that demands deep perception. As shown in Figure~\ref{fig:case}, although vanilla LLaVA-1.5-7B successfully identifies the key visual features (\textit{``bats''}), it tends to rely on its linguistic commonsense knowledge (\textit{i.e.}, \textit{``pumpkin lights''} is a more Halloween-related pattern), leading to an incorrect inference. In contrast, our method enables the LVLM to confidently harness its visual perception capability, accurately identify the key pattern (\textit{``bats''}) and conduct a reasonable inference over the image. Additionally, by comparing the relative attention maps of the two methods, we observe that our method places most attention on the key visual features, demonstrating a more precise and focused attention pattern. Conversely, the vanilla LLaVA-1.5-7B exhibits a more dispersed attention distribution, failing to effectively isolate the key visual features. These results further validate that our method effectively promotes the visual 
 perception and reasoning capabilities of LVLMs.

    
    
     

    
    

    
    
     

    

\begin{table}[t]
    \centering
    \scalebox{0.9}{
    \begin{tabular}{cc|cccc}
    
    \toprule
    \multirow{2}{*}{Rationale} & Direct & \multirow{2}{*}{MMVP} & Wino & {SEED} & \multirow{2}{*}{MM-Vet} \\
    
    & Answer &  & ground & {Bench} &  \\

    \midrule

    \ding{55} & \ding{55} & 20.7 & 25.3 & 66.1 & 31.1 \\
    \ding{51} & \ding{55} & 28.0 & 26.0 & 66.8 & 29.4 \\
    \ding{55} & \ding{51} & 30.7 & 27.0 & 66.3 & 29.2 \\
    \ding{51} & \ding{51} & \textbf{30.7} & \textbf{33.5} & \textbf{67.1} & \textbf{34.1} \\
    
    \bottomrule
    \end{tabular}
    
    }
    \caption{Ablation study of the hybrid loss on LLaVA-1.5-7B.}
    \label{tab:hybrid-formats}
\end{table}

\section{Conclusion}

In this paper, we propose a self-improvement framework that autonomously enhances the visual perception and reasoning capabilities of LVLMs. This framework is grounded in causality-driven visual object completion (CVC), which requires LVLMs to perform visual knowledge-intensive reasoning. We first develop a pipeline for constructing high-causality CVC instances. Then, leveraging trial-and-error learning, we harness the LVLM's inherent reasoning ability to synthesize rationales for each CVC instance and select more challenging ones for self-improvement. Experiments conducted on both challenging specialized tasks and comprehensive benchmarks demonstrate that our method significantly enhances the visual capabilities of LVLMs, particularly in scenarios demanding deep visual perception and reasoning.

\section*{Acknowledgments}

The project was supported by National Key R\&D Program of China (No.2022ZD0160501), Natural Science Foundation of Fujian Province of China (No.2024J011001), and the Public Technology Service Platform Project of Xiamen (No.3502Z20231043). We also thank the reviewers for their insightful comments.


\bibliographystyle{named}
\bibliography{ijcai25}

\clearpage
\appendix
\section*{\centering \LARGE{Appendix}}

\section{Implementation Details}
\label{sec:appendix:implement}

\begin{figure}[h]
    \centering
    \includegraphics[width=\linewidth, trim={0.03in 1.21in 0.03in 0in}, clip]{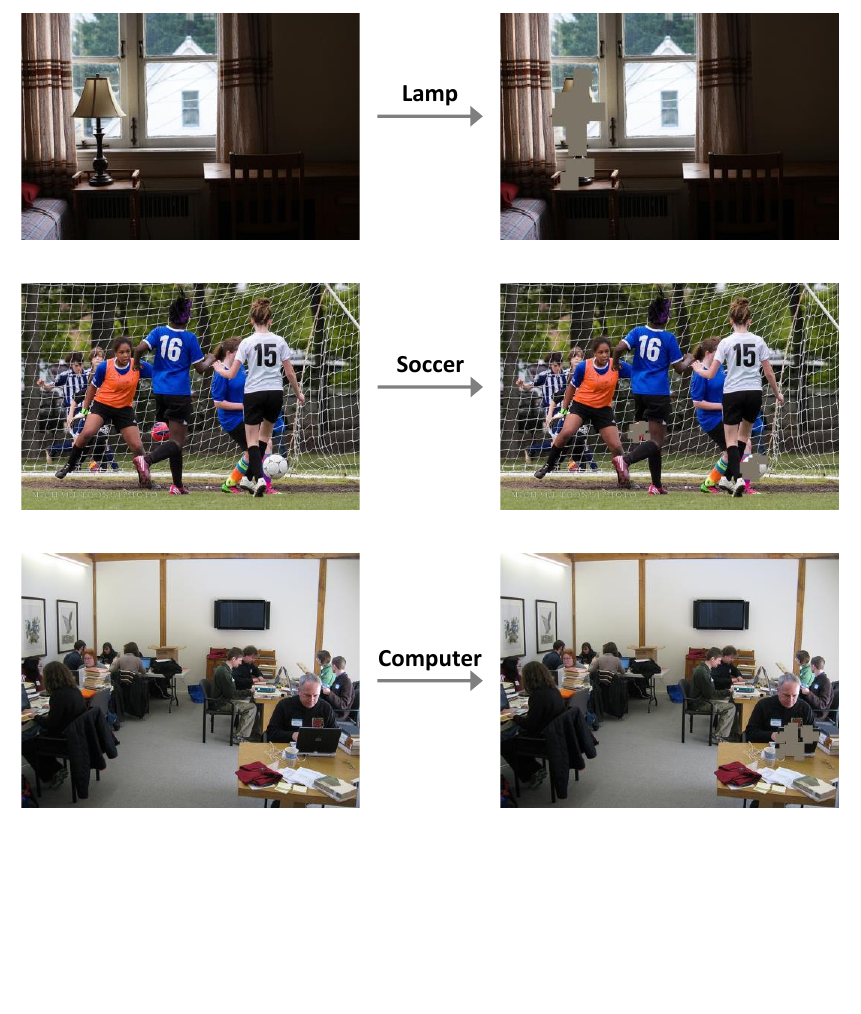}
    \caption{Several examples of occluded images.}
    \label{fig:occluded-images}
\end{figure}

\paragraph{Image Occlusion.}
For each high-causality entity, we first utilize GLIP~\cite{li2022grounded} to ground the corresponding object in the image, generating a bounding box. We then input this bounding box into SAM~\cite{kirillov2023segment} to obtain a mask that precisely delineates the object's region. To seamlessly apply our method across different LVLM architectures, we directly apply occlusion to the segmented pixels. Specifically, we utilize square patches to occlude the object, randomly placing these patches within the object's mask with a small interval. To enhance the occlusion effect on objects of varying sizes, the patch size is set to one-third of the shortest side of the object's bounding box. Additionally, the patch is filled with the mean color of ImageNet~\cite{deng2009imagenet}. We illustrate several examples in Figure~\ref{fig:occluded-images}. The complete implementation is provided in our code repository.
\paragraph{Answer Checking.} In trial and error learning, we need to determine whether the extracted answer aligns with the target answer. To accomplish this, we assess answer correctness through semantic similarity using BGE-M3~\cite{chen2024bge}. The implementation details are available in the provided code.

\section{Prompting Examples}
\label{sec:appendix:prompt}

Our method relies on several prompting examples to elicit the generation from language models through in-context learning. Here, we provide detailed prompting examples used for extracting entities from captions (Table~\ref{tab:entity-extraction-template}), generating specialized instructions for entities (Table~\ref{tab:instruction-generation-template}), and extracting answers from sampled rationales (Table~\ref{tab:answer-extraction-template}).

\begin{table*}[t]
    \centering
    \small
    \noindent\fbox{%
    \begin{minipage}{\dimexpr\linewidth-2\fboxsep-2\fboxrule}
\tt
You are an entity extractor that helps me extract entities from text.
I will provide you with a piece of text, which is a description of an image. Since I need to get the textual descriptions of the entities that appear in the image, I need you to extract entities from the given text as accurately as possible. \\

Here are some definitions of entity:\\
1) Entities are concrete objects that can be recognized in the corresponding image (such as objects, people, places, products).\\
2) The entities to be extracted do not need to be named entities; as long as they are concrete objects, they can be extracted.\\
3) Entities can appear in text in various forms, including single word, noun phrases.\\
4) For abstract concept nouns or noun phrases (such as emotions, states, qualities, sensations, actions), they are not considered as entities.\\

Here are some constraints you need to follow:\\
1) The form of entities does not consider any format other than single word and noun phrases.\\
2) If a noun phrase is not an entity, then the nouns within that phrase cannot be considered as entities either.\\
3) To extract an entity, you first need to identify the complete expression and then remove any modifiers. The text format is as follows: "1. {entity w/ modifiers} -> {entity w/o modifiers}".\\
4) Please do not extract modifiers indicating quantity.\\

Here are some examples:\\

Text: People sunbathing and sitting under umbrellas at a city beach.
Extracted entities:\\
<begin>\\
1. people -> people\\
2. umbrellas -> umbrellas\\
3. city beach -> city beach\\
<end>\\

Text: The image is a collage of various kitchen items, including a clock, pots, pans, a refrigerator, and a chalkboard.\\
Extracted entities:\\
<begin>\\
1. clock -> clock\\
2. pots -> pots\\
3. pans -> pans\\
4. refrigerator -> refrigerator\\
5. chalkboard -> chalkboard\\
<end>\\

Text: A male tennis player hits the ball on a grass court.
Extracted entities:\\
<begin>\\
1. male tennis player -> tennis player\\
2. ball -> ball\\
3. grass court -> grass court\\
<end>\\

Text: A group of people walks down a train platform, with a yellow train stopped nearby.\\
Extracted entities:\\
<begin>\\
1. people -> people\\
2. train platform -> train platform\\
3. yellow train -> train\\
<end>\\

$\cdots$ \\

Text: A motorbike parked on a roadside close to some bush.\\
Extracted entities:\\
<begin>\\
1. motorbike -> motorbike\\
2. bush -> bush\\
<end>\\

    \end{minipage}
}
    \caption{Detailed prompting examples for entity extraction.}
    \label{tab:entity-extraction-template}
\end{table*}
\begin{table*}[t]
    \centering
    \small
    \noindent\fbox{
    \begin{minipage}{\dimexpr\linewidth-2\fboxsep-2\fboxrule}
\tt
You are a question constructor. I will provide you with an entity, which could refer to a specific object, a specific type of object, a place, a location or an occupation.\\
You first need to determine the type of the entity and generate a question according to the type of the given entity.\\

Here are some constraints you need to follow:\\
1) If an object is very common, you can simply refer to it as "object".\\
2) If an object is relatively uncommon, you may need to specify its type in the question.\\
3) If an object's type is difficult to determine, you can simply refer to it as "object".\\
4) Should not reveal the entire entity's information in the generated question.\\

Here are some examples:\\

Entity: sewing machine\\
<begin>\\
Question: In the given image, there is a machine that is heavily occluded by a cluster of gray blocks. Please answer the following question.\\
What kind of machine might the object occluded by a cluster of gray blocks be? Please provide your reasoning process and confirm a unique answer.\\
<end>\\

Entity: refrigerator\\
<begin>\\
Question: In the given image, there is an appliance that is heavily occluded by a cluster of gray blocks. Please answer the following question.\\
What might the appliance occluded by the gray blocks be? Please provide your reasoning process and confirm a unique answer.\\
<end>\\

Entity: fireman\\
<begin>\\
Question: In the given image, there is a person engaged in a certain profession, who is heavily occluded by a cluster of gray blocks. Please answer the following question. \\
What profession might this partially occluded person be engaged in? Please provide your reasoning process and confirm a unique answer.\\
<end>\\

Entity: farm\\
<begin>\\
Question: In the given image, there is a group of person occluded by a cluster of gray blocks. Please answer the following question.\\
What activity might the people occluded by the gray blocks be doing? Please provide your reasoning process and confirm a unique answer.\\
<end>\\

Entity: rider\\
<begin>\\
Question: In the given image, there is a view of a place which is heavily occluded by a cluster of gray blocks. Please answer the following question.\\
The view obscured by the gray blocks could be which place? Please provide your reasoning process and confirm a unique answer.\\
<end>\\

$\cdots$ \\

Entity: couple\\
<begin>\\
Question: In the given image, there is a group of person occluded by a cluster of gray blocks. Please answer the following question.\\
What relationship might the people occluded by the gray blocks have? Please provide your reasoning process and confirm a unique answer.\\
<end>\\

    \end{minipage}
}
    \caption{Detailed prompting examples for instruction generation.}
    \label{tab:instruction-generation-template}
\end{table*}
\begin{table*}[t]
    \centering
    \small
    \noindent\fbox{%
    \begin{minipage}{\dimexpr\linewidth-2\fboxsep-2\fboxrule}
\tt
You are an answer extractor. I will provide you with a text, which is an analysis of reasoning about what an occluded object is. Please extract the confirmed answers from this reasoning process.\\

To extract the answer, here are some constraints you need to follow:\\
1) If the reasoning process does not yield an answer, please determine it as "unknown".\\
2) If the reasoning process expresses uncertainty or cannot be determined, please determine it as "unknown".\\
3) Please directly provide the extracted answer, and do not include your analysis reasons.\\

Here are some examples:\\

Text: Given the heavily occluded object in the image, it's likely to be a piece of food, possibly a meat, which is covered in a generous amount of toppings, such as onions and cheese. However, without seeing more details, the object's precise type and presence of other ingredients cannot be confirmed.\\
<begin>\\
Extracted Answer: meat\\
<end>\\

Text: While looking at the image, I noticed that there's a boy in a red beanie and red jacket sitting in the back seat of the bus. The partially visible object that is heavily blocked by gray blocks might be a window. The initial assumption might be that the boy is looking out of the window, which could be obstructed by the gray blocks. However, since the image also shows another person looking out the window, it's more likely that the gray blocks are there to provide privacy or to separate sections of the bus.\\
<begin>\\
Extracted Answer: window\\
<end>\\

Text: Based on the visual information available in the image, it is impossible to determine the exact nature of the object that is heavily occluded by the gray blocks. However, we can make some educated guesses. One possibility is that the object is a monitor or computer screen, as there is already a computer monitor present in the image. Another possibility is that the object is a lamp or other light source, as there is a lamp present in the image. The gray blocks could also be other random objects, but without more context or visual clues, it is not possible to identify the object with certainty.\\
<begin>\\
Extracted Answer: monitor, computer screen, lamp\\
<end>\\

Text: Based on the image, the heavily occluded object is in a living room area surrounded by brown furniture and plants. Since there is a couch and a chair present in the scene, it's possible that the object is part of the seating arrangement, such as a bench or an ottoman. The gray blocks could be used as an additional seating or as a side table, which could explain the obscured view of the object. The reasoning process involves identifying the living room elements, such as the couch and the chair, and considering the possibility of an additional seating or side table made up of gray blocks.
<begin>\\
Extracted Answer: bench, ottoman\\
<end>\\

Text: The object heavily obscured by the gray blocks might be a white refrigerator or a microwave placed under a cabinet. The gray blocks are blocking any clear view of the object, making it difficult to determine its exact nature. However, these are two common appliances found in kitchens, and their size and color match the description of the heavily blocked object in the image.\\
<begin>\\
Extracted Answer: refrigerator, microwave\\
<end>\\

$\cdots$ \\

Text: The heavily-occluded object in the image is a flat-screen TV mounted on the wall.\\
<begin>\\
Extracted Answer: flat-screen TV\\
<end>\\

    \end{minipage}
}
    \caption{Detailed prompting examples for answer extraction.}
    \label{tab:answer-extraction-template}
\end{table*}

\section{Benchmark Leaderboards}
\label{sec:appendix:leaderboard}
We obtain the results for GPT-4V~\cite{achiam2023gpt} and Gemini Pro~\cite{team2023gemini} on 4 comprehensive benchmarks from their respective official leaderboards. The URL for the leaderboard of each comprehensive benchmark are listed in Table~\ref{tab:leaderboard-url}.

\begin{table*}[t]
    \centering
    \scalebox{1}{
    \begin{tabular}{l|c}
    
    \toprule
    \textbf{Benchmark} & \textbf{Leaderboard URL} \\ 
    
    \midrule
    
    MME~\cite{fu2023mme} & \url{https://github.com/BradyFU/Awesome-Multimodal-Large-Language-Models/tree/Evaluation} \\
    MMBench~\cite{liu2023mmbench} & \url{https://mmbench.opencompass.org.cn/leaderboard} \\
    SEEDBench~\cite{li2023seed} & \url{https://huggingface.co/spaces/AILab-CVC/SEED-Bench_Leaderboard} \\
    MM-Vet~\cite{yu2023mm} & \url{https://paperswithcode.com/sota/visual-question-answering-on-mm-vet}  \\ 
   
    \bottomrule
    \end{tabular}
    }
    \caption{The leaderboard URL of each comprehensive benchmark.}
    \label{tab:leaderboard-url}
\end{table*}

\end{document}